\PassOptionsToPackage{unicode}{hyperref}
\PassOptionsToPackage{hyphens}{url}
\PassOptionsToPackage{dvipsnames,svgnames,x11names}{xcolor}
\documentclass[
]{article}
\usepackage{xcolor}
\usepackage{amsmath,amssymb}
\usepackage{iftex}
\ifPDFTeX
  \usepackage[T1]{fontenc}
  \usepackage[utf8]{inputenc}
  \usepackage{textcomp} 
\else 
  \usepackage{unicode-math} 
  \defaultfontfeatures{Scale=MatchLowercase}
  \defaultfontfeatures[\rmfamily]{Ligatures=TeX,Scale=1}
\fi
\usepackage{lmodern}
\ifPDFTeX\else
\fi
\IfFileExists{upquote.sty}{\usepackage{upquote}}{}
\IfFileExists{microtype.sty}{
  \usepackage[]{microtype}
  \UseMicrotypeSet[protrusion]{basicmath} 
}{}
\makeatletter
\@ifundefined{KOMAClassName}{
  \IfFileExists{parskip.sty}{%
    \usepackage{parskip}
  }{
    \setlength{\parindent}{0pt}
    \setlength{\parskip}{6pt plus 2pt minus 1pt}}
}{
  \KOMAoptions{parskip=half}}
\makeatother
\usepackage{longtable,booktabs,array}
\usepackage{calc} 
\usepackage{etoolbox}
\makeatletter
\patchcmd\longtable{\par}{\if@noskipsec\mbox{}\fi\par}{}{}
\makeatother
\IfFileExists{footnotehyper.sty}{\usepackage{footnotehyper}}{\usepackage{footnote}}
\makesavenoteenv{longtable}
\usepackage{graphicx}
\makeatletter
\newsavebox\pandoc@box
\newcommand*\pandocbounded[1]{
  \sbox\pandoc@box{#1}%
  \Gscale@div\@tempa{\textheight}{\dimexpr\ht\pandoc@box+\dp\pandoc@box\relax}%
  \Gscale@div\@tempb{\linewidth}{\wd\pandoc@box}%
  \ifdim\@tempb\p@<\@tempa\p@\let\@tempa\@tempb\fi
  \ifdim\@tempa\p@<\p@\scalebox{\@tempa}{\usebox\pandoc@box}%
  \else\usebox{\pandoc@box}%
  \fi%
}
\def\fps@figure{htbp}
\makeatother
\providecommand{\tightlist}{%
  \setlength{\itemsep}{0pt}\setlength{\parskip}{0pt}}
\usepackage{arxiv}
\renewcommand{\headeright}{Preprint}
\renewcommand{\undertitle}{Preprint, 23 October 2026}
\renewcommand{\shorttitle}{Metric Scale for Insta360 SfM from the Embedded IMU}
\usepackage{bookmark}
\IfFileExists{xurl.sty}{\usepackage{xurl}}{} 
\hypersetup{
  pdftitle={Metric Scale for Insta360 Structure-from-Motion from the Camera's Embedded IMU},
  pdfauthor={Davide Cuttini, Diego Di Tommaso, Alessio Antoniacomi, Luca
Simonetti},
  colorlinks=true,
  linkcolor={Maroon},
  filecolor={Maroon},
  citecolor={Blue},
  urlcolor={Blue},
  pdfcreator={LaTeX via pandoc}}

\title{Metric Scale for Insta360 Structure-from-Motion from the Camera's
Embedded IMU}
\author{Davide Cuttini \quad Diego Di Tommaso \quad Alessio Antoniacomi$^{\dagger}$ \quad Luca Simonetti$^{\dagger}$ \\ Over The Reality \\ \texttt{\{cuttini, ditommaso, antoniacomi, simonetti\}@ovr.ai} \\ {\small $^{\dagger}$Technical contributors}}
\date{}

\begin{document}
\maketitle
\begin{abstract}
Structure-from-motion (SfM) from the dual-fisheye video of a consumer
360-degree camera recovers geometry only up to a similarity transform;
in the pipeline considered here, metric scale is obtained from
smartphone ARKit or ARCore poses captured alongside the camera. This
paper evaluates a self-contained alternative: the IMU embedded in the
Insta360 X5 and X6, recorded at 1 kHz. The sensor is characterised
(noise density, camera-IMU synchronisation, accelerometer scale factor,
bias stability), the camera-IMU rotation and frame timing are estimated
from the data, and the scale is recovered by a linear visual-inertial
alignment over the SfM poses of a whole survey. The standard
formulation, with gravity and accelerometer bias shared across the
survey, overestimates the scale by 3.4\% on average; estimating each
contiguous segment independently and averaging removes the bias. On nine
surveys with a co-mounted LiDAR scanner the inertial scale deviates from
the LiDAR by 0.81\% in mean absolute error, against 1.94\% for the
smartphone scale; on printed fiducial markers both agree within the 1\%
uncertainty of the target. On the reference trajectories of thirty
sequences of a public benchmark, brought to scales of 1/40, 1 and 40,
the estimator recovers the metric scale with a mean error of $-$0.07\%
and a standard deviation of 1.28\%. On fifty production surveys where
the smartphone is the only reference, the two scales agree to a standard
deviation of 2.3\%, the inertial scale being 0.3\% lower on average; the
difference is uncorrelated with run length, grows with the extent of the
smartphone anchor, and cannot be attributed to either reference. An
ablation over window construction and a comparison with monocular
metric-depth networks complete the study.
\end{abstract}

\section{Introduction}\label{introduction}

Structure-from-motion without a baseline of known length recovers scene
structure and camera trajectory up to a similarity transform: shape and
trajectory are determined, absolute size is not. For a mapping pipeline
that reconstructs scenes from the native dual-fisheye video of an
Insta360 camera, the missing scale must be supplied by an external
measurement. The current solution uses a smartphone carried by the
operator. During the survey the operator captures a burst of photographs
whose poses are metric, from ARKit on iOS or ARCore on Android; the
photographs are registered jointly with the fisheye frames, and a robust
alignment transfers the smartphone scale to the model.

This arrangement has three costs. It requires a second device and a
second capture procedure. It makes the metric reference heterogeneous,
since the two smartphone platforms and their versions behave
differently. And it extrapolates from a small spatial baseline: the
burst is taken within about 15 s over a few metres, so the metric scale
of a survey of several hundred metres rests on a short segment of it,
and any error in the smartphone poses over that segment propagates to
the whole model. Section 5.2 quantifies the consequence: across nine
surveys with a co-mounted LiDAR scanner, the smartphone-anchored models
deviate from the LiDAR by 1.9\% in mean absolute error, with individual
surveys reaching 4.4\%. The smartphone remains a serviceable reference,
and this figure is its measured accuracy rather than a failure; the
question addressed here is whether a sensor already inside the camera
can take its place.

The Insta360 X5 and X6 contain a gyroscope and an accelerometer sampled
at 1000 Hz and written into the recording. The accelerometer measures
specific force in metres per second squared. For a trajectory expressed
in the arbitrary units of an SfM model there is exactly one scale factor
for which its second derivative agrees with the measured accelerations
once gravity is accounted for. Recovering that factor is the
visual-inertial initialisation problem, which is well studied in the
context of real-time SLAM \cite{r16,r17,r21}. The setting here differs
in three respects: the poses are the output of an offline SfM pipeline
rather than a tracker, the alignment is performed in batch over surveys
lasting minutes rather than over the first seconds of motion, and the
frames used by the reconstruction are sparse in time, several per second
at most, while the IMU is continuous.

The contributions of this paper are the following.

\begin{enumerate}
\def\labelenumi{\arabic{enumi}.}
\tightlist
\item
  A characterisation of the embedded IMU of the Insta360 X5 and X6 as
  recorded in the .insv container: noise density, camera-IMU time offset
  and axis mapping, accelerometer scale factor and bias stability.
\item
  A data-driven estimation of the camera-IMU rotation and of the affine
  frame-timing model, together with a diagnostic that detects a wrong
  timing model without ground truth.
\item
  Evidence that the joint linear alignment over a whole survey, with
  shared gravity and accelerometer bias, is biased high by 3.4\% on the
  surveys with external reference, and that a segment-wise estimate
  followed by averaging removes the bias. An ablation over window
  construction shows that the improvement is due to not sharing the
  nuisance parameters and is insensitive to how the survey is divided.
\item
  An evaluation against four independent references: nine surveys with a
  co-mounted LiDAR scanner, one survey with printed fiducial markers of
  known geometry, 62 production surveys with smartphone poses, and the
  reference trajectories of thirty sequences of a public benchmark. On
  the LiDAR surveys the inertial scale is 2.4 times more accurate than
  the smartphone reference.
\item
  A comparison with monocular metric-depth networks as an alternative
  scale source, showing that their scene-dependent bias makes them
  unsuitable as a reference at the accuracy level considered.
\end{enumerate}

The paper is organised as follows. Section 2 reviews related work.
Section 3 describes the method. Section 4 describes the data and the
reference measurements. Section 5 reports the sensor characterisation
and the experimental results. Section 6 discusses the two main findings
and the limitations of the study, and Section 7 concludes.

\section{Related Work}\label{related-work}

\textbf{Visual-inertial initialisation.} The linear relation between
camera displacement, velocity, gravity and preintegrated accelerometer
measurements was formulated in closed form by Martinelli \cite{r16} and
is used to bootstrap visual-inertial SLAM systems, including ORB-SLAM
with inertial support \cite{r17} and VINS-Mono \cite{r21}. Campos et
al.~\cite{r4} refine the linear solution by an inertial-only
maximum-a-posteriori optimisation. Preintegration of the inertial
measurements between camera frames follows Lupton and Sukkarieh
\cite{r15}, placed on the manifold by Forster et al.~\cite{r8}. These
systems initialise over a few seconds of motion and then track online.
The present work applies the same linear formulation offline, in batch
over a whole survey, to poses produced by an SfM pipeline; the
differences that arise at that time scale, in particular the behaviour
of the shared gravity and bias parameters, are the subject of Section
5.6 and Section 6.1.

\textbf{Camera-IMU calibration.} The joint estimation of the camera-IMU
extrinsic rotation and time offset is addressed by the continuous-time
calibration of Furgale et al.~\cite{r9}, which requires a calibration
target. Here the rotation is obtained by a hand-eye formulation on
rotation vectors solved with Kabsch's method \cite{r13}, and the time
model is restricted to a two-parameter affine map estimated from the
survey itself, with no target.

\textbf{Sensor characterisation.} Noise density is estimated from the
amplitude spectral density with Welch's method \cite{r28}. Bias
instability is conventionally characterised by the Allan variance of a
long static recording \cite{r12,r6}.

\textbf{Alternative scale references.} Metric depth networks such as
Metric3D \cite{r30}, ZoeDepth \cite{r2}, Depth Anything \cite{r29} and
MoGe \cite{r27} predict absolute depth from a single image and could in
principle provide scale without inertial data; Section 5.9 evaluates two
of them. Fiducial markers of known size \cite{r10} provide a local
metric reference. LiDAR SLAM of the LOAM family \cite{r31} provides a
metric map whose registration to the reconstruction, by a closed-form
initialisation \cite{r26} followed by iterative closest point
\cite{r1,r22}, yields the scale of the model.

\textbf{Reconstruction.} The SfM models used throughout are produced by
a standard incremental pipeline in the COLMAP family \cite{r24}, with
learned features and retrieval-based matching in the manner of Sarlin et
al.~\cite{r23} and an equidistant fisheye camera model \cite{r14}.

\section{Method}\label{method}

\subsection{Problem statement and
notation}\label{problem-statement-and-notation}

The input is an SfM model with camera poses
\((\mathbf{R}_{WC,k}, \mathbf{C}_k)\) for frames \(k = 1, \dots, N\),
where \(\mathbf{R}_{WC,k}\) is the rotation from the camera frame \(C\)
to the model frame \(W\) and \(\mathbf{C}_k\) is the camera centre in
model units, together with gyroscope and accelerometer measurements
\(\boldsymbol{\omega}(\tau)\) and \(\mathbf{a}(\tau)\) in the IMU body
frame \(B\) at 1 kHz. The unknowns are the scale \(s\) that converts
model units to metres, the gravity vector
\(\mathbf{g} \in \mathbb{R}^3\) in the model frame, the accelerometer
bias \(\mathbf{b}_a \in \mathbb{R}^3\) and one velocity \(\mathbf{v}_k\)
per frame. Two further quantities must be determined from the data
before the alignment can be formed: the rotation \(\mathbf{R}_{CB}\)
between camera and IMU frames, and the timing model \(t_k = a\,i_k + b\)
that maps the video frame index \(i_k\) of frame \(k\) onto the IMU
clock.

\subsection{Sensor data and timing
model}\label{sensor-data-and-timing-model}

The camera records two HEVC streams of 3840 by 3840 pixels, one per
lens, without stitching. The IMU record is read from the container
metadata with the telemetry-parser library \cite{r11}. Gyroscope and
accelerometer are sampled at 1000 Hz with no gaps on recordings of up to
27 min. The X6 stores integer sensor counts; the X5 stores
floating-point values that the firmware has low-pass filtered. Files
exported by the manufacturer's desktop software lose the metadata.

The frames used by the reconstruction are taken from the video at a
fixed nominal interval, so the time of frame \(i\) on the IMU clock is
\(t_i = a\,i + b\). The nominal value of the interval \(a\) is not
accurate enough for the alignment: an error of 0.1\% displaces the
video-to-IMU correspondence by 0.6 s over a ten-minute survey, more than
a frame interval, and the scale estimate fails. Both timing constants
are therefore estimated from the data (Section 3.3); the nominal
interval serves only as the starting point of the search, and on one of
the fifty surveys in Section 5.5 the interval that fits the data differs
from the nominal one by 0.16\%, enough to displace the correspondence by
0.9 s over a nine-minute survey. The offset search of Section 3.3 must
therefore cover a range of at least \textpm{}1 s once the interval is
refined.

\subsection{Camera-IMU rotation and temporal
calibration}\label{camera-imu-rotation-and-temporal-calibration}

For consecutive frames the relative rotation from SfM,
\(\mathbf{R}_{C}^{k,k+1} = \mathbf{R}_{WC,k}^{\top}\mathbf{R}_{WC,k+1}\),
must equal the relative rotation integrated from the gyroscope over the
same interval, \(\mathbf{R}_{B}^{k,k+1}\), expressed in the camera
frame:

\[
\mathbf{R}_{C}^{k,k+1} = \mathbf{R}_{CB}\,\mathbf{R}_{B}^{k,k+1}\,\mathbf{R}_{CB}^{\top}. \tag{1}
\]

On the corresponding rotation vectors this reduces to
\(\mathbf{r}_C = \mathbf{R}_{CB}\,\mathbf{r}_B\), and
\(\mathbf{R}_{CB}\) is obtained by Kabsch's method over all frame pairs.
The timing constants \(a\) and \(b\) are found by minimising the median
angular residual of Equation (1). The search is performed in two stages:
a coarse grid over candidate intervals, then a fine grid of
\(\pm 0.4\%\) around the best candidate with the offset re-centred,
followed by a local simplex refinement. The two-stage search is
necessary. A single unconstrained simplex started from the nominal
interval converges only when the motion is smooth; on brisk walking it
returns the starting value, leaving a residual of 3 to 4 degrees per
interval and a scale estimate in error by tens of percent. With the
correct interval the residual is 0.05 to 0.18 degrees per interval on
every survey in this study, and the recovered \(\mathbf{R}_{CB}\) is
identical across independent surveys, as expected for a rigid sensor
assembly.

A wrong timing model is detectable without ground truth. The optimal
time offset \(b\) is recomputed on the first and the last quarter of the
survey; if the interval \(a\) is wrong, the two estimates differ by a
drift proportional to the error. On the surveys of Section 5 the drift
is below 10 ms, whereas the nominal interval produces 530 ms on a
ten-minute survey. The estimator reports failure when the drift exceeds
60 ms.

\subsection{Linear visual-inertial
alignment}\label{linear-visual-inertial-alignment}

For each interval \([t_k, t_{k+1}]\) between consecutive frames the
accelerometer is integrated at 1 kHz in the model frame. The body
orientation \(\mathbf{R}_{WB}(\tau)\) is initialised from the SfM
orientation of frame \(k\) through the camera-IMU rotation,
\(\mathbf{R}_{WB}(t_k) = \mathbf{R}_{WC,k}\,\mathbf{R}_{CB}\), and
propagated by the gyroscope. Four quantities are accumulated over the
interval, with \(\Delta t_k = t_{k+1} - t_k\):

\[
\Delta\mathbf{p}_k = \int_{t_k}^{t_{k+1}}\!\!\int_{t_k}^{\tau} \mathbf{R}_{WB}(\sigma)\,\mathbf{a}(\sigma)\,d\sigma\,d\tau,
\qquad
\Delta\mathbf{v}_k = \int_{t_k}^{t_{k+1}} \mathbf{R}_{WB}(\tau)\,\mathbf{a}(\tau)\,d\tau, \tag{2}
\]

\[
\mathbf{J}_k = \int_{t_k}^{t_{k+1}}\!\!\int_{t_k}^{\tau} \mathbf{R}_{WB}(\sigma)\,d\sigma\,d\tau,
\qquad
\mathbf{K}_k = \int_{t_k}^{t_{k+1}} \mathbf{R}_{WB}(\tau)\,d\tau . \tag{3}
\]

This is the preintegration of Lupton and Sukkarieh \cite{r15}, in the
form of Forster et al.~\cite{r8}; only the first-order dependence on the
bias is retained, and the matrices \(\mathbf{J}_k\) and \(\mathbf{K}_k\)
make the dependence on \(\mathbf{b}_a\) linear. Each interval
contributes a position equation and a velocity equation:

\[
s\,(\mathbf{C}_{k+1} - \mathbf{C}_k) - \mathbf{v}_k\,\Delta t_k - \tfrac{1}{2}\,\mathbf{g}\,\Delta t_k^{2} + \mathbf{J}_k\,\mathbf{b}_a = \Delta\mathbf{p}_k , \tag{4}
\]

\[
\mathbf{v}_{k+1} - \mathbf{v}_k - \mathbf{g}\,\Delta t_k + \mathbf{K}_k\,\mathbf{b}_a = \Delta\mathbf{v}_k . \tag{5}
\]

Equations (4) and (5) are linear in
\((s, \mathbf{g}, \mathbf{b}_a, \mathbf{v}_1, \dots, \mathbf{v}_N)\) and
are solved in closed form through sparse normal equations. This is the
linear initialisation of Martinelli \cite{r16}, Mur-Artal and Tardos
\cite{r17} and Qin et al.~\cite{r21}, applied in batch to a whole survey
and without the subsequent non-linear or inertial-only refinement of
Campos et al.~\cite{r4}. Velocities are chained only within a contiguous
run of frames, defined as a maximal sequence of frames separated by no
more than 1.2 s; runs are interrupted where the pipeline has discarded
frames for insufficient sharpness.

\subsection{Segment-wise estimation}\label{segment-wise-estimation}

The remaining design choice is how the runs are combined. The joint
formulation solves all runs together as one system in which \(s\),
\(\mathbf{g}\) and \(\mathbf{b}_a\) are shared, on the grounds that
gravity and sensor bias are properties of the survey rather than of the
run. Measured against external ground truth on ten surveys, this
formulation overestimates the scale by 3.39\% on average, with the same
sign on nine of the ten (Section 5.6). The joint solution is not an
average of the per-run solutions and is not constrained to lie among
them; any violation of the assumption that gravity and bias are constant
across runs recorded minutes apart is absorbed by \(s\), because gravity
enters each position equation as
\(\tfrac{1}{2}\mathbf{g}\,\Delta t_k^2\), a term of the same order as
the displacement itself.

The estimator adopted here therefore solves each run separately, with
its own \(s\), \(\mathbf{g}\), \(\mathbf{b}_a\) and velocities, and
reports the mean of the per-run scales. Runs shorter than eight
intervals, and runs whose recovered gravity magnitude lies outside 9.3
to 10.3 m\,s$^{-2}$, are discarded. When fewer than four usable runs
remain, as in a recording from which no frame was discarded, the runs
are cut into consecutive windows of 25 intervals before the per-window
solve; this fallback does not trigger on any survey of this study. On
the same ten surveys the segment-wise estimate reduces the mean absolute
error from 3.43\% to 0.85\% and removes the systematic sign. The
dispersion of the per-run values provides the uncertainty of the mean
directly.

Two biases of opposite sign bound the per-run estimate. Runs that are
too short are biased low, because noise in the SfM camera centres enters
Equation (4) as an error in the regressor and attenuates the estimated
scale; the ablation of Section 5.6 measures this attenuation at $-$0.6\%
for sliding windows of 15 intervals and approximately zero for 30. Long
windows are biased high for the reason given above, by +0.9\% at 60
intervals and +3.4\% for the whole survey. Contiguous runs, whose
typical length in this pipeline is 15 to 60 intervals, lie close to the
point where both effects vanish.

\subsection{Quality criteria}\label{quality-criteria}

Several quantities are computed to make the estimate verifiable, all of
them from the per-run solutions rather than from the joint one. Gravity
is left unconstrained in every run: its magnitude is not needed for the
scale, but a correct alignment must recover it, and a run whose gravity
magnitude falls outside 9.3 to 10.3 m\,s$^{-2}$ indicates a wrong timing
model, a wrong camera-IMU rotation or a defective reconstruction and is
discarded; a survey in which such runs hold more than half of the
intervals is reported as failed. The gravity direction used to level the
model is the interval-weighted mean of the per-run gravity vectors. The
dispersion of the per-run scales is reported with the estimate; its
standard deviation across runs is 1 to 3\% on the surveys of Section 5,
so the standard error of the mean over five to thirteen runs is 0.3 to
1.5\%. A survey whose standard error exceeds 2.5\%, or which yields
fewer than four usable runs, is reported as failed. The per-run scales
of the first and second half of the trajectory are averaged separately;
a disagreement above 6\% indicates internal scale drift in the model,
and the per-run values identify segments where the scale is weakly
observable, which occurs when the operator moves at constant velocity
and the accelerometer measures little beyond gravity. The joint solution
and its bootstrap dispersion are still computed and reported as
diagnostics, since a large gap between the joint solution and the
per-run mean indicates that gravity or bias is not constant over the
survey, but they do not enter the acceptance decision. Intervals whose
residual exceeds five times the median are removed before a second
solve. The second solve isolates segments in which the SfM itself is
misregistered: on one survey with 19 m discontinuities in the trajectory
the estimator does not converge after outlier removal and reports
failure rather than a scale.

\section{Experimental Setup}\label{experimental-setup}

\subsection{Reconstruction pipeline and
datasets}\label{reconstruction-pipeline-and-datasets}

All poses used in this study are taken unchanged from the production
reconstruction pipeline: COLMAP-format models built from the native
dual-fisheye frames with learned features, retrieval-based matching and
a two-lens rig constraint in bundle adjustment, that is, the standard
components of current large-scale SfM. No reconstruction was recomputed
for this work. The estimator reads the frames of one lens of the rig;
lens 1 is used on every survey, a choice fixed before any result was
read, and lens 0 is reported as a robustness check. All results are
computed with the public implementation of the estimator at its default
settings \cite{r18}. Each registered frame carries its video frame
index, so frame times follow from it, and frames discarded by the
sharpness filter leave gaps without renumbering.

Five sets of surveys are used (Table 1). The LiDAR set consists of nine
surveys acquired with a 3DMakerPro Eagle LiDAR scanner carried on the
same pole as an Insta360 X5: one of 112 m from a first campaign (L1) and
eight of 101 to 241 m from a second campaign (L2 to L9), lasting 100 to
228 s. L1 carries ARKit poses from an iPhone; L2 to L9 carry ARCore
poses from Android phones. The fiducial set is one survey of 159 m and
203 s containing two printed A3 sheets with ArUco markers. The first
production set consists of twelve surveys of the current pipeline
generation for which an ARKit-anchored scale and an intact
reconstruction were available, including one indoor recording
reprocessed with the current pipeline for this study; they span indoor
and outdoor scenes, with durations of 53 to 600 s and trajectories of 49
to 775 m. The second production set consists of fifty surveys, already
scaled by the smartphone registration, of 246 to 708 s and 192 to 1254 m
each, 33 km in total; this set is publicly released as OverMaps360-50
\cite{r20},
\url{https://huggingface.co/datasets/OverTheReality/OverMaps360_50}. The
nine LiDAR surveys are released with their recordings, reconstructions,
LiDAR clouds and registrations \cite{r19}. A fifth set, used in Sections
5.7 and 5.8 only, consists of the thirty sequences of the public
Hilti-Trimble-Oxford benchmark \cite{r5}, recorded indoors on a
construction site with an Insta360 ONE RS 1-inch camera and its embedded
IMU at 1 kHz, 69 to 276 s and 72 to 352 m each, with a camera trajectory
measured by a rigidly attached LiDAR-inertial system. For this set no
reconstruction is used: the reference trajectory itself is supplied to
the estimator.

\begin{longtable}[]{@{}
  >{\raggedright\arraybackslash}p{(\linewidth - 8\tabcolsep) * \real{0.2000}}
  >{\raggedright\arraybackslash}p{(\linewidth - 8\tabcolsep) * \real{0.2000}}
  >{\raggedright\arraybackslash}p{(\linewidth - 8\tabcolsep) * \real{0.2000}}
  >{\raggedright\arraybackslash}p{(\linewidth - 8\tabcolsep) * \real{0.2000}}
  >{\raggedright\arraybackslash}p{(\linewidth - 8\tabcolsep) * \real{0.2000}}@{}}
\caption{Survey sets used in the evaluation. The models of the first
four sets are dual-fisheye rig reconstructions from the production
pipeline; for the public benchmark the input to the estimator is the
reference trajectory.}\tabularnewline
\toprule\noalign{}
\begin{minipage}[b]{\linewidth}\raggedright
set
\end{minipage} & \begin{minipage}[b]{\linewidth}\raggedright
surveys
\end{minipage} & \begin{minipage}[b]{\linewidth}\raggedright
duration
\end{minipage} & \begin{minipage}[b]{\linewidth}\raggedright
path length
\end{minipage} & \begin{minipage}[b]{\linewidth}\raggedright
reference
\end{minipage} \\
\midrule\noalign{}
\endfirsthead
\toprule\noalign{}
\begin{minipage}[b]{\linewidth}\raggedright
set
\end{minipage} & \begin{minipage}[b]{\linewidth}\raggedright
surveys
\end{minipage} & \begin{minipage}[b]{\linewidth}\raggedright
duration
\end{minipage} & \begin{minipage}[b]{\linewidth}\raggedright
path length
\end{minipage} & \begin{minipage}[b]{\linewidth}\raggedright
reference
\end{minipage} \\
\midrule\noalign{}
\endhead
\bottomrule\noalign{}
\endlastfoot
LiDAR & 9 & 100 to 228 s & 101 to 241 m & co-mounted LiDAR; ARKit (1)
and ARCore (8) \\
Fiducial & 1 & 203 s & 159 m & ArUco markers; ARKit \\
Production A & 12 & 53 to 600 s & 49 to 775 m & ARKit \\
Production B & 50 & 246 to 708 s & 192 to 1254 m (33 km in total) &
ARKit \\
Public benchmark & 30 & 69 to 276 s & 72 to 352 m (5.5 km in total) &
LiDAR-inertial trajectory \\
\end{longtable}

Four further recordings were examined and excluded: one whose
reconstruction contains 19 m discontinuities, on which the estimator
reports failure; one bundle-adjustment experiment in non-metric units;
and two 80 s recordings without smartphone photographs, hence without a
reference. Recordings from the previous pipeline generation, whose
perspective views were cut from a stabilised equirectangular projection,
cannot be used, because stabilisation decouples the image orientation
from the gyroscope.

\subsection{Reference measurements}\label{reference-measurements}

\textbf{LiDAR.} The scanner builds its map with LiDAR SLAM of the LOAM
family. For each survey the LiDAR cloud and the pipeline reconstruction
were aligned by a closed-form trajectory initialisation followed by a
scaled iterative-closest-point cascade on the dense clouds, which
returns the factor converting the reconstruction into LiDAR metres.
Since the reconstruction is anchored to the smartphone, that factor is
also the error of the smartphone reference. The scanner's specified
accuracy is 2 cm at 10 m, 3 cm at 20 m and 5 cm at 40 m. This is a
per-point figure and does not limit a scale comparison: the factor is
recovered from millions of points along a whole trajectory, so the
random part averages out, while the systematic part of a time-of-flight
range is set by a crystal oscillator. The LiDAR map may drift in shape
over long paths, but every distance within it is measured in metres
directly, which is the quantity the reconstruction lacks.

\textbf{Fiducial markers.} Two A3 sheets, each carrying two ArUco
markers of the $4\times4$ dictionary with a nominal side of 140 mm and a
nominal centre-to-centre distance of 255 mm, were placed in the scene.
Marker corners were detected with sub-pixel refinement, restricted to
views within 50 degrees of the optical axis, unprojected with the
model's fisheye intrinsics and triangulated across 23 to 28 views per
marker. Views beyond 50 degrees were excluded because the fisheye
distortion model of the reconstruction biases the triangulated marker
size there by several percent.

\textbf{Smartphone.} The smartphone reference is the scale obtained by
the robust registration of the ARKit or ARCore poses of the photograph
burst to the fisheye reconstruction, as described in Section 1.

\section{Results}\label{results}

\subsection{Sensor characterisation}\label{sensor-characterisation}

\textbf{Noise density.} No static recording was available at the outset,
and a moving recording contains the operator's motion superimposed on
sensor noise. The two are separable in frequency: walking and hand
motion occupy the band below 50 Hz, so the band from 100 Hz to the
Nyquist frequency of 500 Hz contains sensor noise only. The amplitude
spectral density of the gyroscope was estimated with Welch's method over
80 s of hand-held recording (Figure 1). The X6 spectrum is flat above
100 Hz at about 0.006 degrees per second per square root hertz, white
noise at the level of a good consumer MEMS device. The X5 spectrum
continues to fall with frequency, identifying the firmware low-pass
filter. The accelerometer shows an internal low-pass filter as well,
with an upper bound of 1 to 5 mm\,s$^{-2}$\,Hz$^{-1/2}$ on its noise
density. The sensor noise is therefore compatible with visual-inertial
alignment, and the X5 values are filtered rather than raw.

\begin{figure}
\centering
\includegraphics[width=1\linewidth,height=\textheight,keepaspectratio]{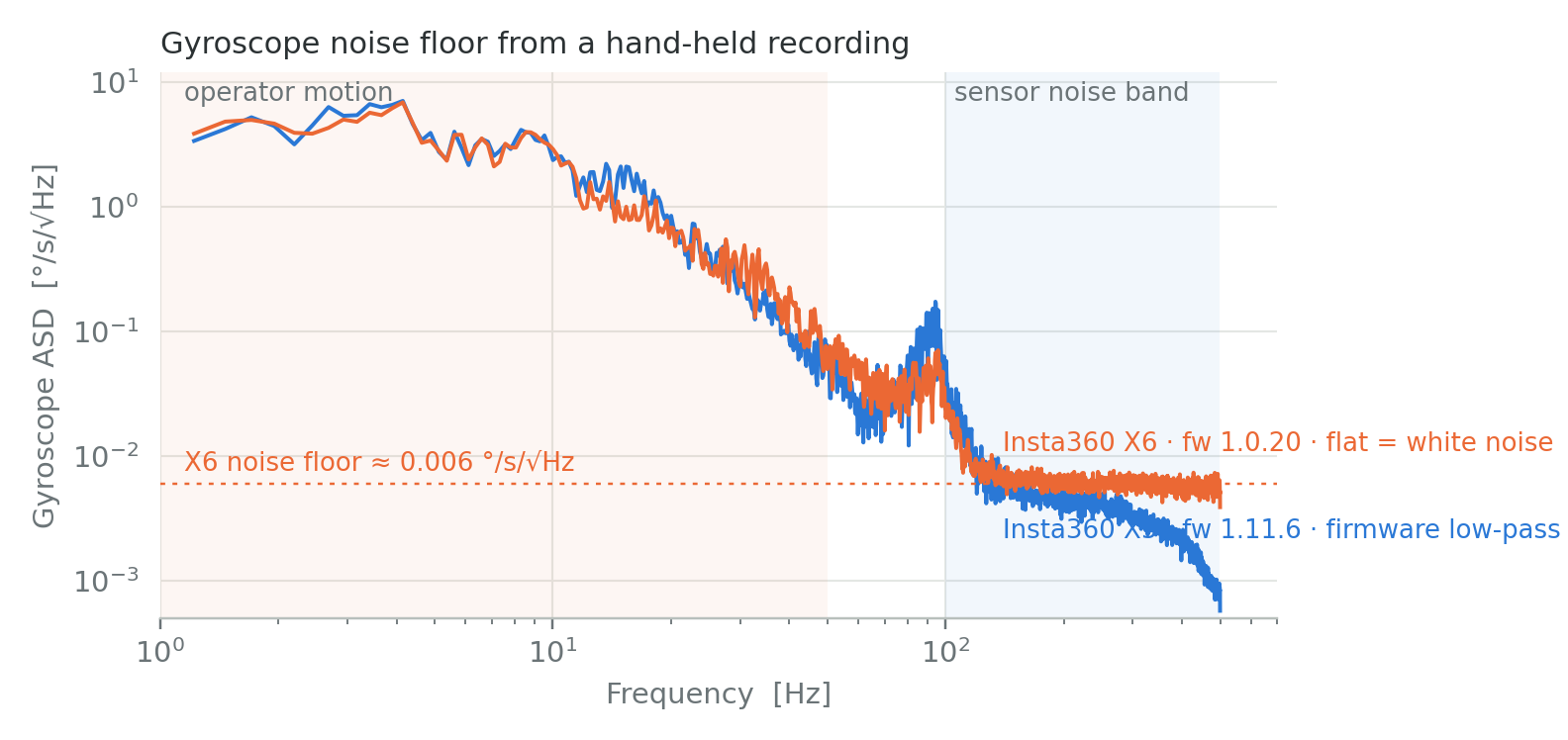}
\caption{Amplitude spectral density of the gyroscope y axis for the
Insta360 X5 (blue) and X6 (orange) over 80 s of hand-held recording,
estimated with Welch's method (segment length 4096). The band below 50
Hz is dominated by operator motion; above 100 Hz only sensor noise
remains. The X6 is flat (white noise); the X5 is attenuated by a
firmware low-pass filter.}
\end{figure}

\textbf{Synchronisation and axis mapping.} Integrating the accelerometer
over a frame interval presumes that frame timestamps and IMU timestamps
refer to the same clock; a residual offset would shift every integration
window and bias the scale. To test this without a calibration target,
the camera rotation was estimated from the video alone and compared with
the gyroscope. Shi-Tomasi corners \cite{r25} were tracked between
consecutive frames with pyramidal Lucas-Kanade optical flow \cite{r3}
and a forward-backward consistency check; the tracks were unprojected
with an equidistant fisheye model, and the inter-frame rotation was
estimated by Kabsch alignment inside a RANSAC loop \cite{r7} under a
pure-rotation assumption. The resulting angular rate was compared with
the gyroscope rate averaged over the same frame interval, for gyroscope
time shifts from $-$150 to +150 ms in 1 ms steps and for all signed
permutations of the three axes, retaining the combination with the
highest mean absolute correlation. On a hand-held sequence with angular
rates up to 480 degrees per second the mean correlation is 0.993 and the
maximum lies at $-$2 ms (Figure 2): the IMU timestamps in the file
already refer to the video clock. The correlation curve is broad because
each visual sample averages the rotation over a 42 ms frame interval, so
the offset is bounded to about \textpm{}5 ms rather than resolved to the
millisecond. The axis mapping is identical across firmware versions
1.0.6 and 1.11.6, and the rear lens is related to the front lens by a
180-degree rotation about the vertical image axis.

\begin{figure}
\centering
\includegraphics[width=1\linewidth,height=\textheight,keepaspectratio]{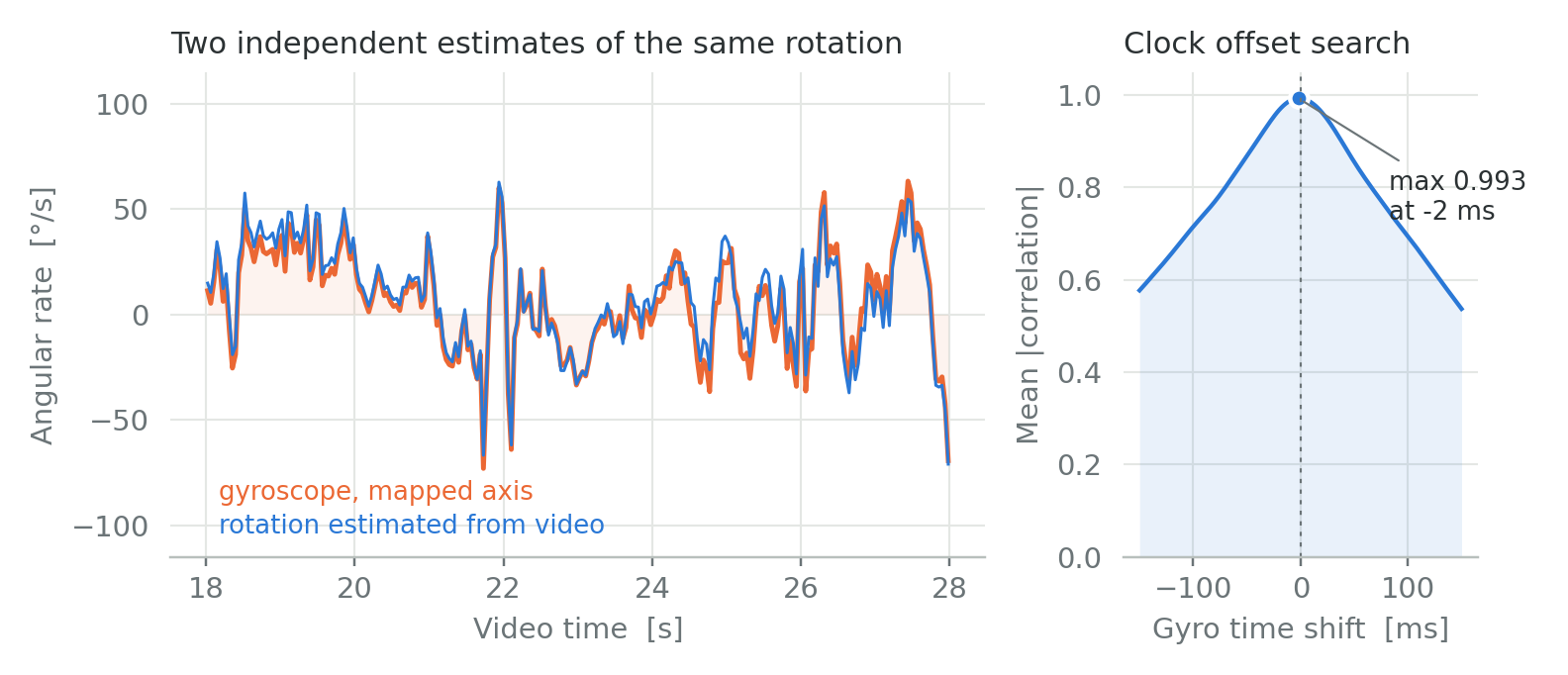}
\caption{Left: angular rate about the dominant axis over 10 s of a
hand-held sequence with fast rotations, from video-only rotation
estimation (blue) and from the gyroscope after axis mapping (orange).
Right: mean absolute correlation between the two as a function of the
time shift applied to the gyroscope; the maximum lies at $-$2 ms, within
the \textpm{}5 ms resolution of the test, showing that the IMU
timestamps stored in the file are aligned to the video.}
\end{figure}

\textbf{Accelerometer scale factor and bias stability.} Two properties
of the accelerometer bear directly on the scale: whether its scale
factor is correct, and how far its zero moves during a survey. Both were
obtained from an eleven-minute recording with the camera resting on a
table. The gyroscope confirms that the camera is stationary, with the
norm of the angular rate below 0.44 degrees per second at the 99th
percentile. The mean magnitude of the measured specific force is 9.8092
m\,s$^{-2}$, against a local gravity of 9.806 m\,s$^{-2}$ at the
latitude of the recording; the accelerometer scale factor is therefore
correct to 0.03\%, which excludes a sensor gain error as an explanation
for any scale discrepancy. The bias drifts monotonically with the shape
of a thermal warm-up, fastest in the first five minutes and flattening
afterwards. Averaged in ten-second blocks, the x axis moves from
$-$0.376 to $-$0.408 m\,s$^{-2}$ and the z axis from $-$0.108 to
$-$0.132 m\,s$^{-2}$, a vector excursion of 0.042 m\,s$^{-2}$ over ten
and a half minutes; on a survey of three minutes the expected excursion
is about 0.013 m\,s$^{-2}$. This is an order of magnitude below the
value that would be required to explain the bias of the joint
formulation (Section 6.1). The gyroscope bias over the same recording
has a magnitude of 0.18 degrees per second with an excursion of 0.30
degrees per second, which amounts to 0.06 degrees over one frame
interval and is negligible for the hand-eye estimate. The recording is
from a third X5 unit, not one of the two used in the surveys; it
establishes the order of magnitude of both effects rather than the exact
figures for those units.

\subsection{Co-mounted LiDAR}\label{co-mounted-lidar}

Table 2 reports the scale error of the inertial estimate and of the
smartphone reference relative to the LiDAR on the nine surveys; a
positive error denotes a scale larger than the LiDAR. The inertial scale
has a mean error of +0.07\%, a mean absolute error of 0.81\% and a worst
case of 2.11\%. The smartphone reference has a mean error of $-$0.41\%,
a mean absolute error of 1.94\% and a worst case of 4.36\%. Neither
reference shows a systematic sign, as expected of two unbiased
references, but the inertial scale is the closer of the two on seven of
the nine surveys and is more accurate by a factor of 2.4 overall (Figure
4). The smartphone column mixes ARKit and ARCore, which is
representative of the fleet and is one of the costs noted in Section 1.
The dispersion of the per-run inertial estimates within each survey is
1.1\% to 3.2\%, giving a standard error of the mean between 0.5\% and
1.4\%, consistent with the scatter observed against the LiDAR.

\begin{longtable}[]{@{}lllll@{}}
\caption{Scale error relative to the co-mounted LiDAR on nine surveys,
with path lengths in LiDAR metres, for the smartphone reference that
anchors the reconstruction and for the inertial estimate. Mean absolute
error: 1.94\% (smartphone), 0.81\% (IMU). With lens 0 instead of lens 1
the IMU mean absolute error is 0.76\%.}\tabularnewline
\toprule\noalign{}
survey & campaign & path & smartphone error & IMU error \\
\midrule\noalign{}
\endfirsthead
\toprule\noalign{}
survey & campaign & path & smartphone error & IMU error \\
\midrule\noalign{}
\endhead
\bottomrule\noalign{}
\endlastfoot
L1 & first & 112 m & +2.92\% & $-$2.11\% \\
L2 & second & 140 m & $-$1.18\% & +2.06\% \\
L3 & second & 105 m & $-$4.36\% & +0.02\% \\
L4 & second & 106 m & +0.63\% & +1.27\% \\
L5 & second & 101 m & $-$2.79\% & +0.39\% \\
L6 & second & 233 m & +0.68\% & $-$0.10\% \\
L7 & second & 225 m & $-$2.24\% & +0.25\% \\
L8 & second & 210 m & +0.76\% & $-$0.01\% \\
L9 & second & 241 m & +1.87\% & $-$1.10\% \\
\end{longtable}

One survey, L5, was flagged by an earlier version of the quality
criteria that compared the two halves of the trajectory under the joint
solution, although its scale is accurate to 0.39\%; Section 6.3
describes why the criteria were moved to the per-run quantities.

\begin{figure}
\centering
\includegraphics[width=1\linewidth,height=\textheight,keepaspectratio]{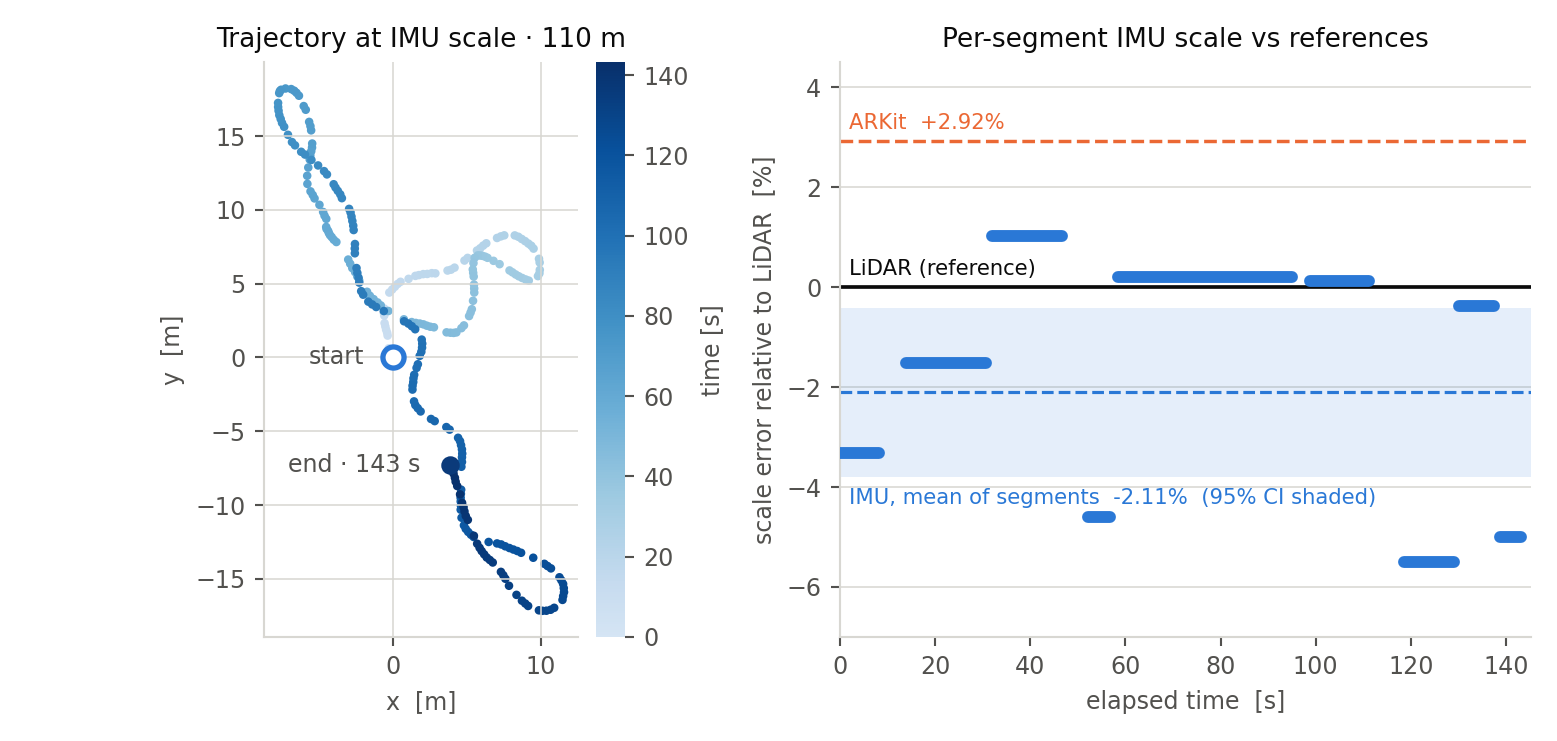}
\caption{LiDAR survey L1, with the co-mounted LiDAR as reference. Left:
camera trajectory from the production SfM model, scaled by the IMU
estimate and aligned to the recovered gravity direction, coloured by
elapsed time (110 m at the IMU scale, 112 m in LiDAR metres, over 143 s
of registered frames). Right: scale error relative to the LiDAR (solid
line at zero) of the inertial estimate on each contiguous segment
(bars), of the mean of the segments with its 95\% confidence interval
(dashed blue line and band, $-$2.11\%) and of the ARKit reference
(dashed orange, +2.92\%), as in Table 2.}
\end{figure}

\begin{figure}
\centering
\includegraphics[width=1\linewidth,height=\textheight,keepaspectratio]{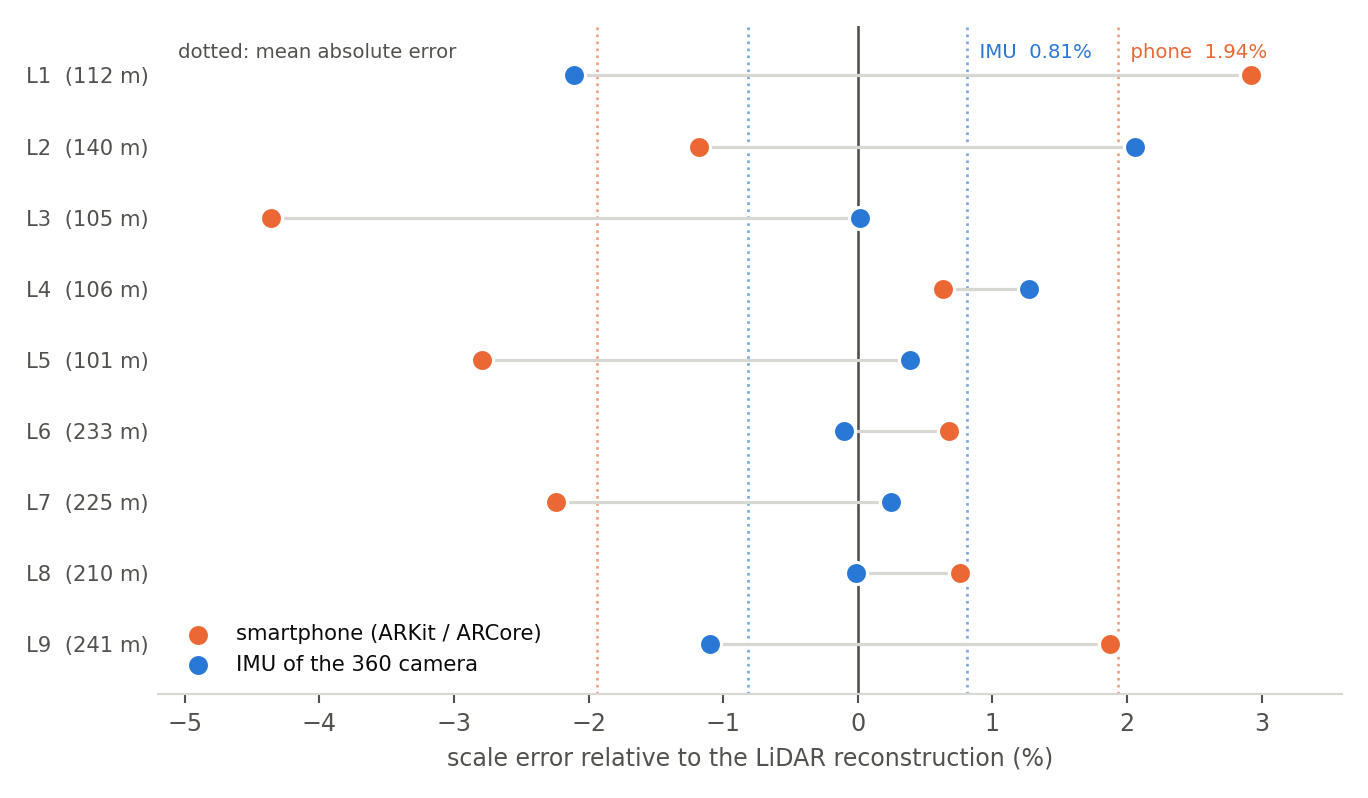}
\caption{Scale error relative to the LiDAR reconstruction on the nine
co-mounted surveys, for the inertial scale and for the smartphone
reference that anchors the reconstruction. Each point is one survey; the
dotted lines mark the mean absolute error of each source. The monocular
depth networks of Section 5.9, evaluated on survey L1 only, fall outside
the range of this plot.}
\end{figure}

\subsection{Fiducial markers}\label{fiducial-markers}

The marker reference itself carries a 1\% uncertainty. Marker sides
measured in model units imply a scale of 1.011 to 1.021, whereas centre
distances imply 0.997 to 0.998; the discrepancy is systematic on both
sheets and corresponds to an inward bias of the corner localisation of
1.5\% to 2\% at a marker size of about 130 pixels, so the centre-based
estimate is preferred. Rectifying the sheet through the marker
homography shows a paper width of 300 to 301 mm against the A3 nominal
of 297 mm, that is, the printed content is about 1\% smaller than the
layout. After this correction the marker-derived scale lies between
0.987 and 1.005. The inertial estimate on this survey is 0.989 and the
ARKit scale is 1.000; both fall inside the marker interval (Figure 5).

\begin{figure}
\centering
\includegraphics[width=1\linewidth,height=\textheight,keepaspectratio]{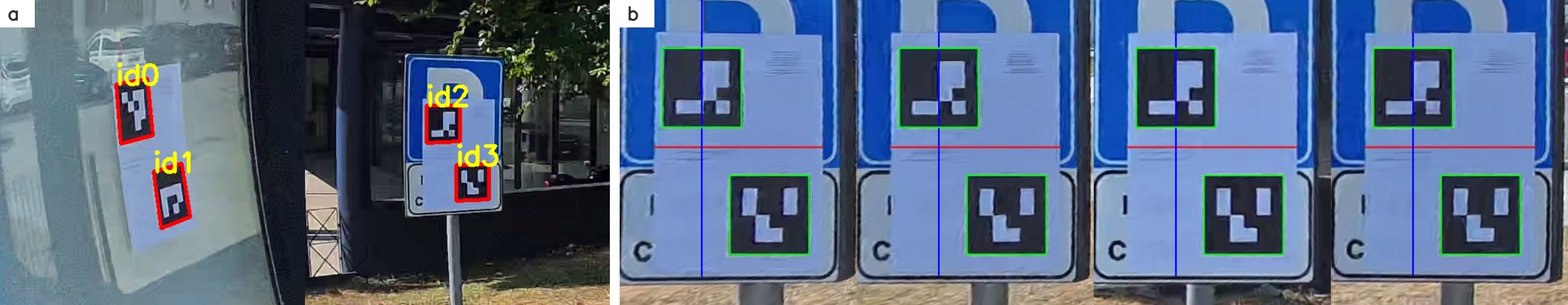}
\caption{(a) The two ArUco sheets as seen in the fisheye frames, one
viewed obliquely through glass (ids 0 and 1, excluded) and one on a
rigid signboard (ids 2 and 3). (b) The signboard sheet rectified through
the homography defined by the eight detected corners, with the nominal
140 mm marker outlines (green) overlaid; the black squares fall slightly
inside the nominal outlines, consistent with a print scale about 1\%
below nominal.}
\end{figure}

\subsection{Production set A: twelve
surveys}\label{production-set-a-twelve-surveys}

All twelve surveys were processed with the same settings and without
manual intervention. The ratio between the inertial scale and the ARKit
scale has a mean of 0.991, a median of 0.994 and a standard deviation of
3.0\%, with extremes of 0.947 and 1.035 (Figure 6, left). The lowest
ratio belongs to the reprocessed indoor recording (A12), in which the
smartphone photographs cover an area of only 2 by 1.4 m; a monocular
depth estimate on the same model agrees with the inertial scale (0.956
against ARKit). The internal uncertainty of the inertial estimate, a
median disagreement of 1.5\% between trajectory halves, accounts for
part of the dispersion; the remainder is attributable to the ARKit
reference, in line with the 2.9\% error of that reference on LiDAR
survey L1. No systematic offset between the two references is resolved
on this sample; Section 5.5 examines the offset on a larger one. The
recovered gravity magnitude is 9.77 \textpm{} 0.08 m\,s$^{-2}$ across
the surveys.

The values in this section were computed with the joint estimator of
Section 3.5 and have not been recomputed for the whole set, because nine
of the twelve reconstructions are no longer available. Three were re-run
with the segment-wise estimator: the ratio to the smartphone scale
changes by $-$0.3, +0.5 and $-$2.7 percentage points, the last being
A12, the survey with the 2 by 1.4 m smartphone coverage. The dispersion
quoted above is therefore an upper bound; the part attributable to the
estimator is removed by the segment-wise formulation, while the part
attributable to the smartphone reference is not.

\begin{figure}
\centering
\includegraphics[width=1\linewidth,height=\textheight,keepaspectratio]{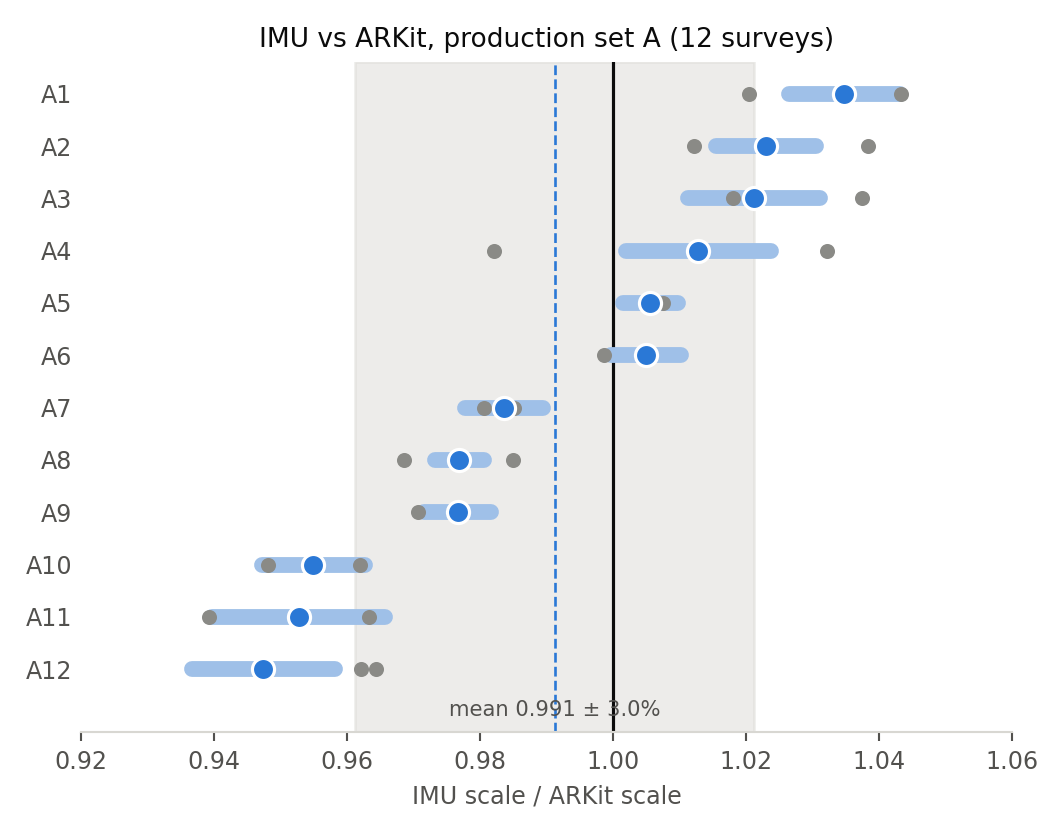}
\caption{Ratio between the IMU-derived scale and the ARKit-derived scale
for the twelve surveys of production set A (A1 to A12), sorted by value.
Filled circles are full-trajectory estimates, shaded bars the
\(\pm1\sigma\) bootstrap interval, small grey circles the estimates from
the first and second halves of each trajectory; the grey band is the
mean \textpm{} one standard deviation across surveys.}
\end{figure}

\subsection{Production set B: fifty
surveys}\label{production-set-b-fifty-surveys}

The estimator was run unchanged on the fifty surveys of production set
B, and all fifty produce an estimate that passes the criteria of Section
3.6. The inertial scale agrees with the ARKit reference to a standard
deviation of 2.3\%, with a median internal uncertainty of 0.36\%, a
median per-run dispersion of 2.0\%, a median recovered gravity of 9.79
m\,s$^{-2}$ and a median hand-eye residual of 0.08 degrees. The
estimator is therefore as well conditioned on this set as on the surveys
with ground truth; the dispersion of its agreement with the smartphone
scale is comparable to that of set A (3.0\%).

The mean ratio is 0.997 and the median 0.994, so the inertial scale lies
0.3\% below the smartphone scale on average, and 31 of the fifty surveys
fall below unity, with extremes of 0.956 and 1.076 (Figure 7, left).
Neither reference is ground truth on this set. The smartphone scale
carries a measured error of 1.9\% against LiDAR (Section 5.2), and the
inertial scale 0.8\%; a difference of 0.3\% between them is therefore
within what the two errors allow, and it does not place the two
references in contradiction. It does show that they are not
interchangeable survey by survey, and that without an external
instrument on the same set the sign of the offset cannot be assigned to
either.

The first factor examined is the window-length dependence of Section
5.6: the recordings of set B are longer than those of set A, and their
contiguous runs could be long enough to inherit the upward bias of the
joint solution. The correlation between mean run length and the ratio is
$-$0.15 across the set (Figure 7, right), with mean run lengths between
22 and 105 intervals; the correlation with survey duration is 0.00 and
with path length +0.07. The second factor is the lens. With lens 0
instead of lens 1 all fifty surveys pass as well, and the mean ratio is
0.991, 0.6\% lower; on the LiDAR surveys the two lenses differ by a
similar amount in either direction and have a similar accuracy (Table
2). The choice of lens therefore moves the mean ratio on this set by an
amount comparable to its offset from unity, and the origin of the
difference between the two lenses has not been isolated.

The third factor is the extent of the smartphone anchor. The photograph
bursts of set B contain 42 to 400 photographs spanning 3.2 to 29.0 m,
with a median span of 6.3 m. The ratio correlates with that span (r =
+0.52; Spearman 0.50) and with the number of photographs (r = +0.53). On
the 42 surveys whose anchor spans less than 12 m the inertial scale lies
0.8\% below the smartphone scale (median $-$0.8\%, standard deviation
1.7\%, 29 of the 42 below unity), and within this group the correlation
with the span is +0.31; the eight surveys with an anchor of 16 to 29 m
and 180 to 400 photographs have a mean of +2.4\%, ranging from $-$1.6\%
to +7.6\%. The dependence is carried largely by these eight surveys, and
one of them, at +7.6\%, weighs heavily on the correlation. The extreme
values are not explained by the diagnostics of the inertial estimate:
the survey at $-$4.4\% has 47 photographs spanning 3.7 m, a hand-eye
residual of 0.08 degrees, a recovered gravity of 9.86 m\,s$^{-2}$, a
per-run dispersion of 3.2\% and the two trajectory halves agreeing to
0.9\%. No third reference exists on this set, so the difference between
the two scales cannot be attributed with the data in hand; Section 6.2
discusses what can be concluded without one.

The set also exercises the joint solution beyond the ten surveys of
Section 5.6. On 34 of the fifty surveys it lies above the segment-wise
mean, by 0.4\% in median, a milder version of the 3.4\% bias measured
there; on three it fails outright, returning scales between 0.000 and
0.064 while the segment-wise estimates on the same surveys remain tight,
with 2.5\% dispersion and 0.4\% to 0.7\% uncertainty on the mean, and on
three more its recovered gravity magnitude falls outside 9.3 to 10.3
m\,s$^{-2}$ while the per-run gravity is normal. Under the per-run
criteria of Section 3.6 all fifty surveys pass, with a median
disagreement between the two halves of 0.7\% and a maximum of 2.6\%.

\begin{figure}
\centering
\includegraphics[width=1\linewidth,height=\textheight,keepaspectratio]{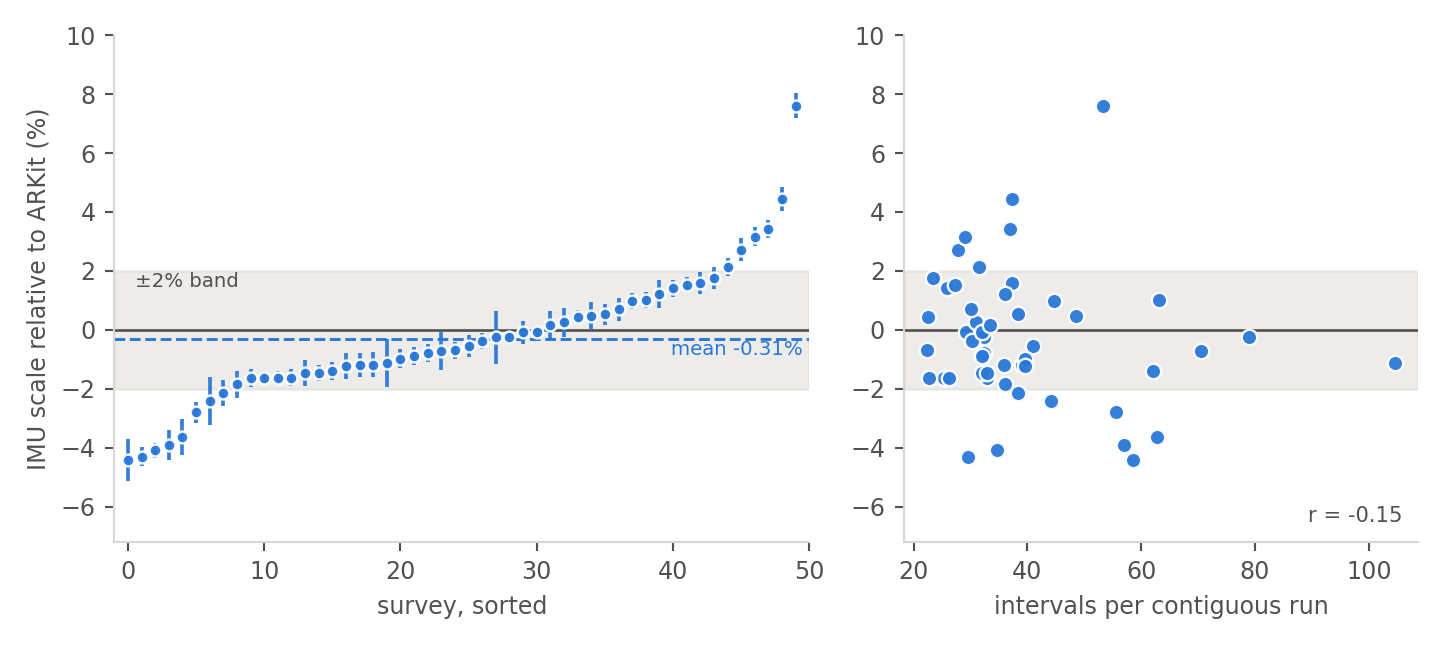}
\caption{Production set B, fifty surveys without external reference.
Left: ratio between the IMU-derived scale and the ARKit-derived scale on
the fifty surveys, sorted by value, with the uncertainty on the mean of
the per-run estimates as error bar, the \textpm{}2\% band shaded and the
set mean dashed. Right: the same ratio against the mean length of the
contiguous runs it is estimated on; the runs span 22 to 105 intervals
and the correlation is $-$0.15, so the window-length bias of Section 5.6
does not account for the difference.}
\end{figure}

\subsection{Ablation: window
construction}\label{ablation-window-construction}

This section separates the two decisions involved in the segment-wise
estimate, how the survey is divided into windows and how the window
estimates are combined, and measures each against the ten surveys with
external ground truth (the nine LiDAR surveys and the fiducial survey).
Three window constructions were compared. \emph{Contiguous runs} uses
the natural segmentation, in which a run ends wherever the pipeline has
discarded frames. \emph{Fixed windows} cuts each run into consecutive
blocks of \(N\) intervals. \emph{Sliding windows} takes every window of
\(N\) intervals at stride one, so that windows overlap and each interval
is counted many times. A stride variant interpolates between the last
two, taking windows of 30 intervals at a stride of 15. Every variant
estimates each window independently and reports the mean.

\begin{longtable}[]{@{}
  >{\raggedright\arraybackslash}p{(\linewidth - 8\tabcolsep) * \real{0.2000}}
  >{\raggedright\arraybackslash}p{(\linewidth - 8\tabcolsep) * \real{0.2000}}
  >{\raggedright\arraybackslash}p{(\linewidth - 8\tabcolsep) * \real{0.2000}}
  >{\raggedright\arraybackslash}p{(\linewidth - 8\tabcolsep) * \real{0.2000}}
  >{\raggedright\arraybackslash}p{(\linewidth - 8\tabcolsep) * \real{0.2000}}@{}}
\caption{Scale error against external ground truth on the ten surveys
with LiDAR or fiducial reference, for each window construction.
``Surveys covered'' is the number of surveys on which the construction
yields at least one valid window.}\tabularnewline
\toprule\noalign{}
\begin{minipage}[b]{\linewidth}\raggedright
window construction
\end{minipage} & \begin{minipage}[b]{\linewidth}\raggedright
surveys covered
\end{minipage} & \begin{minipage}[b]{\linewidth}\raggedright
mean error
\end{minipage} & \begin{minipage}[b]{\linewidth}\raggedright
mean absolute error
\end{minipage} & \begin{minipage}[b]{\linewidth}\raggedright
worst
\end{minipage} \\
\midrule\noalign{}
\endfirsthead
\toprule\noalign{}
\begin{minipage}[b]{\linewidth}\raggedright
window construction
\end{minipage} & \begin{minipage}[b]{\linewidth}\raggedright
surveys covered
\end{minipage} & \begin{minipage}[b]{\linewidth}\raggedright
mean error
\end{minipage} & \begin{minipage}[b]{\linewidth}\raggedright
mean absolute error
\end{minipage} & \begin{minipage}[b]{\linewidth}\raggedright
worst
\end{minipage} \\
\midrule\noalign{}
\endhead
\bottomrule\noalign{}
\endlastfoot
whole survey as one system & 10 & +3.39\% & 3.43\% & 6.05\% \\
contiguous runs & 10 & $-$0.05\% & 0.85\% & 2.11\% \\
30 intervals, stride 15 & 10 & $-$0.20\% & 0.89\% & 1.95\% \\
fixed, 30 intervals & 10 & $-$0.32\% & 0.90\% & 2.23\% \\
fixed, 50 intervals & 10 & $-$0.24\% & 0.90\% & 1.91\% \\
fixed, 25 intervals & 10 & $-$0.44\% & 0.95\% & 1.52\% \\
sliding, 30 intervals & 10 & $-$0.00\% & 0.97\% & 2.09\% \\
sliding, 60 intervals & 10 & +0.86\% & 1.17\% & 4.56\% \\
sliding, 15 intervals & 10 & $-$0.63\% & 1.22\% & 2.42\% \\
fixed, 15 intervals & 10 & $-$0.54\% & 1.39\% & 3.01\% \\
\end{longtable}

Three observations follow from Table 3 and Figure 8. First, every
windowed variant improves on the single joint system by a factor between
2.5 and 4, so the gain of the segment-wise estimate comes from not
sharing gravity and bias across the survey rather than from any
particular segmentation. Second, the mean absolute error of the windowed
constructions ranges from 0.85\% to 1.39\%, and the contiguous runs
adopted in Section 3.5 have the lowest, by a few hundredths of a point
over the next constructions; the method is insensitive to this choice,
and the contiguous runs are the one construction with no length to set.
Third, window length does matter, in the direction predicted in Section
3.5: sliding windows of 15 intervals give a mean error of $-$0.6\%, 30
intervals give approximately zero, and 60 intervals give +0.9\% as the
window begins to inherit the bias of the joint solution. Overlap has
little effect: sliding windows at stride one produce five to fifteen
times as many estimates as non-overlapping ones and change the mean
absolute error by less than 0.1 percentage points.

\begin{figure}
\centering
\includegraphics[width=1\linewidth,height=\textheight,keepaspectratio]{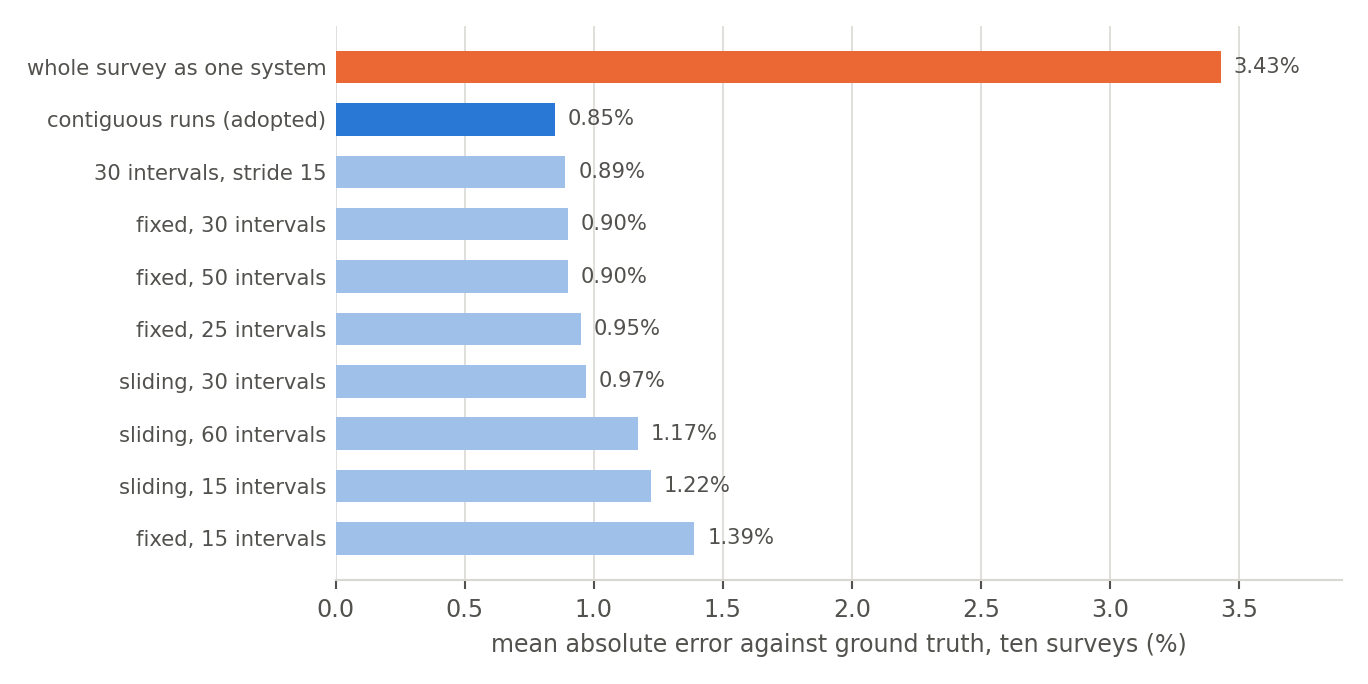}
\caption{Mean absolute error against ground truth on the ten surveys
with external reference, for each window construction of Table 3.
Orange: the single joint system; blue: the contiguous-run construction
adopted here. Every windowed variant improves on the joint system by a
factor between 2.5 and 4, and all of them lie within 0.6 percentage
points of each other.}
\end{figure}

\subsection{Invariance to the input
model}\label{invariance-to-the-input-model}

The scale of the input model is unknown by assumption and its
orientation is arbitrary. Gross per-run outliers are therefore rejected
relative to the median of the per-run scales, never against an absolute
range, and the joint solution is not used as a reference because it can
fail on its own (Section 6.1). To confirm that the estimator exploits
neither the scale nor the orientation of the input, each of the thirty
reference trajectories of the public benchmark (Section 5.8) was
transformed by a similarity with a scale factor of 1/40, 1 and 40, a
random rotation and a random translation, and supplied to the estimator.
On every sequence the three recovered scales are identical once divided
by the applied factor, to within the printed precision of 0.0001\%, and
every run passes the quality criteria. The input model may therefore be
supplied in any units and any orientation.

\subsection{Reference trajectories of a public
benchmark}\label{reference-trajectories-of-a-public-benchmark}

To evaluate the estimator independently of any reconstruction, it was
run on the reference trajectories of the public benchmark of Table 1,
decimated to 3 Hz and brought to an arbitrary scale as in Section 5.7.
The recovered scale has a mean error of $-$0.07\% across the thirty
sequences, a median of $-$0.24\%, a standard deviation of 1.28\% and a
worst case of 3.2\%; 18 sequences lie within 1\% and 28 within 2\%, and
all thirty pass the quality criteria of Section 3.6. On the twenty-five
sequences used for ranking by the benchmark the mean error is $-$0.26\%
and the mean absolute error 0.85\%. On trajectories without
reconstruction error the estimator is therefore unbiased, and its
dispersion per sequence on recordings of one to five minutes is about
1\%.

\subsection{Monocular metric depth as an alternative scale
source}\label{monocular-metric-depth-as-an-alternative-scale-source}

Metric depth networks were evaluated as a scale source that requires no
inertial data, using MoGe-2 (ViT-L) and MoGe-3 (ViT-g). Fisheye frames
were rectified to 90-degree pinhole views, the network was run with the
known field of view, and for each frame the median ratio between
predicted depth and the depth of the reprojected sparse SfM points gives
a scale estimate; the median over frames is the survey estimate.

On LiDAR survey L1 (A7 in set A), an outdoor open area of 40 by 25 m
evaluated on all 290 frames, MoGe-2 deviates from the LiDAR by $-$3.2\%
and MoGe-3 by $-$18.8\%, with a per-frame scatter of 8\% to 9\%. Within
a single frame the predicted depth is proportional to the SfM depth to
about 1\%; the error lies in the global scale the network assigns to the
scene, not in its shape. To test whether the outdoor character of the
site explains the discrepancy, both networks were run on ten surveys of
set A, four indoor and six outdoor, using 30 frames per survey and the
production poses and sparse points as geometric reference (Table 4,
Figure 9). Because these models are ARKit-scaled, the ratio between
depth-derived and model scale is the network's bias with respect to the
ARKit reference; the inertial ratio is listed for comparison.

\begin{longtable}[]{@{}
  >{\raggedright\arraybackslash}p{(\linewidth - 8\tabcolsep) * \real{0.2000}}
  >{\raggedright\arraybackslash}p{(\linewidth - 8\tabcolsep) * \real{0.2000}}
  >{\raggedright\arraybackslash}p{(\linewidth - 8\tabcolsep) * \real{0.2000}}
  >{\raggedright\arraybackslash}p{(\linewidth - 8\tabcolsep) * \real{0.2000}}
  >{\raggedright\arraybackslash}p{(\linewidth - 8\tabcolsep) * \real{0.2000}}@{}}
\caption{Scale from monocular metric depth relative to the ARKit-scaled
production model on ten surveys, 30 frames each, with the inertial ratio
for comparison. The inertial ratio is the segment-wise estimate of
Sections 5.2 and 5.3 for A7 and A5, and the joint estimate of Section
5.4 for the other surveys, whose models are no longer
available.}\tabularnewline
\toprule\noalign{}
\begin{minipage}[b]{\linewidth}\raggedright
survey
\end{minipage} & \begin{minipage}[b]{\linewidth}\raggedright
scene
\end{minipage} & \begin{minipage}[b]{\linewidth}\raggedright
MoGe-2 / ARKit
\end{minipage} & \begin{minipage}[b]{\linewidth}\raggedright
MoGe-3 / ARKit
\end{minipage} & \begin{minipage}[b]{\linewidth}\raggedright
IMU / ARKit
\end{minipage} \\
\midrule\noalign{}
\endfirsthead
\toprule\noalign{}
\begin{minipage}[b]{\linewidth}\raggedright
survey
\end{minipage} & \begin{minipage}[b]{\linewidth}\raggedright
scene
\end{minipage} & \begin{minipage}[b]{\linewidth}\raggedright
MoGe-2 / ARKit
\end{minipage} & \begin{minipage}[b]{\linewidth}\raggedright
MoGe-3 / ARKit
\end{minipage} & \begin{minipage}[b]{\linewidth}\raggedright
IMU / ARKit
\end{minipage} \\
\midrule\noalign{}
\endhead
\bottomrule\noalign{}
\endlastfoot
A10 & indoor & 0.832 ($-$16.8\%) & 0.839 ($-$16.1\%) & 0.955 \\
A12 & indoor & 0.956 ($-$4.4\%) & 0.889 ($-$11.1\%) & 0.947 \\
A1 & indoor & 0.993 ($-$0.7\%) & 0.974 ($-$2.6\%) & 1.035 \\
A3 & indoor & 0.999 ($-$0.1\%) & 1.004 (+0.4\%) & 1.021 \\
A11 & outdoor & 0.773 ($-$22.7\%) & 0.924 ($-$7.6\%) & 0.953 \\
A4 & outdoor & 0.834 ($-$16.6\%) & 1.178 (+17.8\%) & 1.013 \\
A6 & outdoor & 0.931 ($-$6.9\%) & 0.923 ($-$7.7\%) & 1.005 \\
A7 (LiDAR L1) & outdoor & 0.941 ($-$5.9\%) & 0.789 ($-$21.1\%) &
0.951 \\
A5 (fiducial) & outdoor & 0.977 ($-$2.3\%) & 0.871 ($-$12.9\%) &
0.989 \\
A2 & outdoor & 1.086 (+8.6\%) & 0.945 ($-$5.5\%) & 1.023 \\
\end{longtable}

Outdoor surveys are worse on average, with a mean absolute bias of 11\%
for MoGe-2 and 12\% for MoGe-3 against 5\% and 8\% indoors, but the
split is not clean: two indoor surveys (A1, A3) are recovered within 1\%
by both networks, while a third indoor survey (A10) is off by 16\% with
both. The two networks also disagree with each other on the same scenes,
most visibly on A4, where MoGe-2 reads $-$17\% and MoGe-3 +18\%, and on
A7, $-$6\% against $-$21\%. The newer and larger MoGe-3 is not more
accurate. The controlling variable appears to be the proximity of the
scene to the networks' training distribution rather than the
indoor/outdoor distinction as such. On the same ten surveys the inertial
scale stays within 6\% of ARKit in every case. Since the two networks
are trained by the same group on largely the same data, their agreement
on a given scene is not evidence of correctness. A depth-based estimate
can serve as a coarse consistency check on the inertial one, but not as
the metric reference: its bias depends on the scene and, at the level of
accuracy at stake here, exceeds the disagreement it would be asked to
resolve.

\begin{figure}
\centering
\includegraphics[width=1\linewidth,height=\textheight,keepaspectratio]{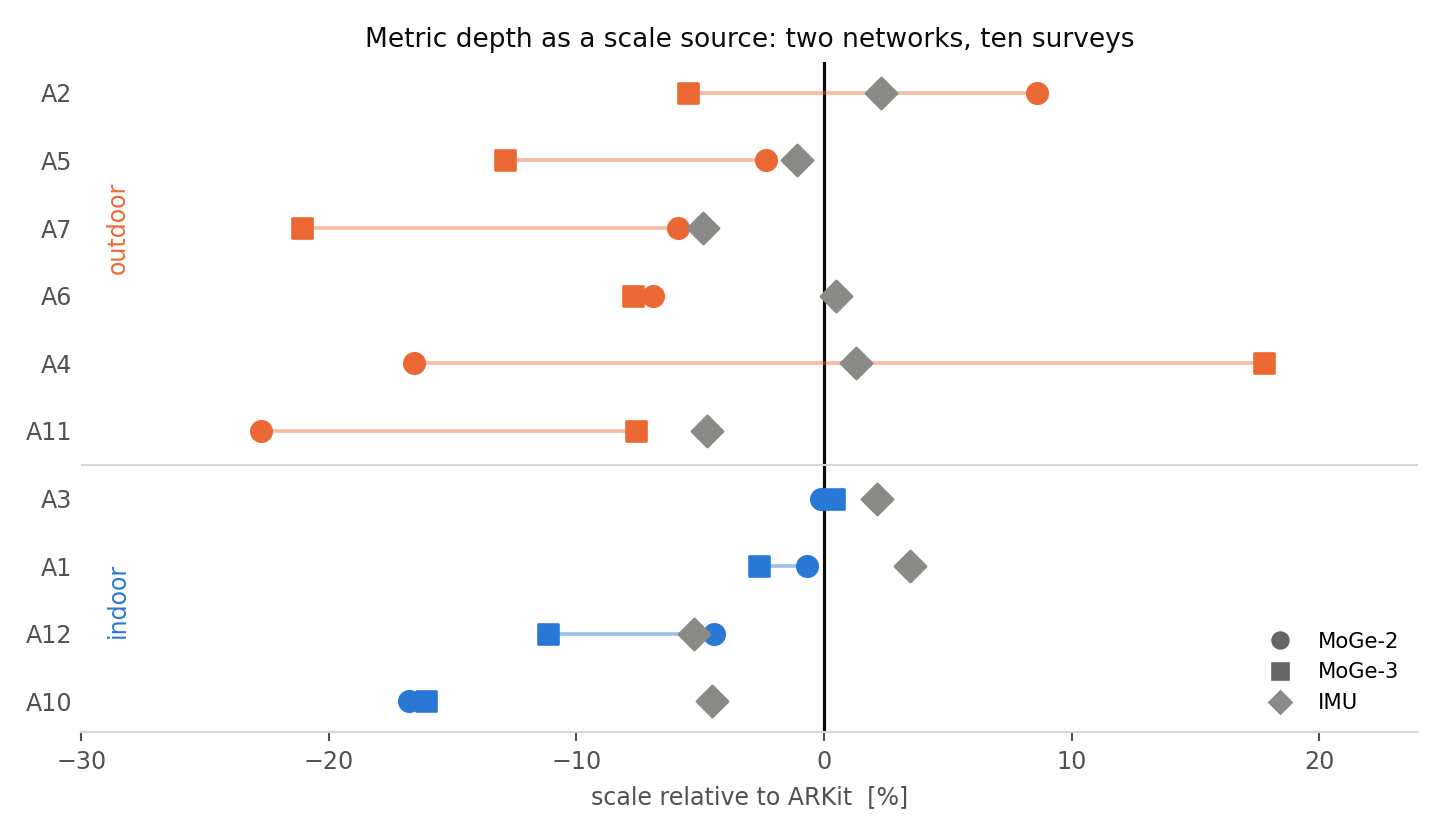}
\caption{Scale bias of MoGe-2 (circles) and MoGe-3 (squares) metric
depth relative to the ARKit-scaled production model on ten surveys of
set A, four indoor (blue) and six outdoor (orange), from 30 frames each;
grey diamonds mark the inertial scale. Outdoor surveys are worse on
average, but one indoor survey fails with both networks and the two
networks disagree by up to 34\% on the same scene.}
\end{figure}

\section{Discussion}\label{discussion}

\subsection{Bias of the joint
formulation}\label{bias-of-the-joint-formulation}

The largest single effect in the method is the choice between a joint
solution with shared nuisance parameters and a segment-wise solution.
Against external ground truth the joint solution is biased high by
3.39\% with consistent sign (Table 3), and on the fifty surveys of set B
it lies above the segment-wise mean on 34 and fails on three. On the
reference trajectories of the public benchmark, which carry no
reconstruction error, the joint solution is unbiased ($-$0.06\% on
average), which places the source of the bias in the reconstructions
rather than in the sensor. The static recording of Section 5.1 bounds
the accelerometer bias drift at 0.042 m\,s$^{-2}$ over ten minutes, an
order of magnitude below what would be required to produce the observed
shift through the bias term alone, so bias drift is excluded as the
cause. The remaining candidate is the gravity term. If the effective
gravity direction in the model frame is not constant along the survey,
for instance because of residual orientation drift in the
reconstruction, the shared-gravity assumption is violated, and because
\(\tfrac{1}{2}\mathbf{g}\,\Delta t_k^2\) is of the same order as the
displacement in Equation (4), the violation is absorbed by \(s\). The
ablation of Section 5.6 is consistent with this reading: the bias grows
with window length, and every construction that estimates gravity
locally removes it regardless of how the windows are placed. The
specific mechanism that makes the effective gravity vary along a survey
has not been isolated in this study. The per-run estimate is bounded
from below by regressor-noise attenuation on short windows and from
above by the shared-gravity bias on long ones, and contiguous runs of 15
to 60 intervals lie near the crossover.

\subsection{Offset on the production sample and the absence of ground
truth}\label{offset-on-the-production-sample-and-the-absence-of-ground-truth}

On fifty production surveys the inertial scale lies 0.3\% below the
smartphone scale on average, on 31 of the fifty. The difference is not
correlated with run length, survey duration or path length, and it is of
the same size as the difference between the two lenses of the rig
(0.6\%). It does vary with the spatial extent of the smartphone anchor:
on the 42 surveys anchored over less than 12 m the inertial scale is
0.8\% lower, while on the eight anchored over 16 to 29 m it is 2.4\%
higher. The difference between the two references is therefore not a
constant property of either, and on average it is smaller than the error
the smartphone reference carries against LiDAR, so the two references
are not in contradiction. The point to retain is that on production
surveys no ground truth exists: the smartphone scale is itself an
estimate with a measured error of 1.9\% and a worst case of 4.4\%, and
an agreement of 2.3\% between two independent estimates of that quality
is what one expects when both are approximately right. Taken together,
the LiDAR, fiducial and smartphone comparisons bound the error of the
inertial scale at below 1\% in mean absolute error, 2.1\% in the worst
case, where an instrument reading metres is available and at 3 to 4\% in
the worst case against the smartphone, with no configuration in which
the inertial estimate disagrees with every other reference. The source
of the residual difference on set B, and of its dependence on the
anchor, can only be assigned by a third, metric measurement on the same
scenes, which is left to future work.

\subsection{Quality criteria}\label{quality-criteria-1}

The quality criteria were originally designed for the joint estimator,
and three of them, the recovered gravity magnitude, the bootstrap
dispersion and the half-versus-half comparison, were read from the joint
solution. On an earlier sample of fifty production surveys that version
rejected six whose segment-wise estimates are unremarkable, and on the
LiDAR set it rejected one survey whose scale is accurate to 0.39\%; in
every such case the joint solution had either collapsed or recovered a
gravity magnitude far from 9.81 m\,s$^{-2}$ while the per-run solutions
were normal. The criteria were therefore restated on the per-run
quantities, as described in Section 3.6: gravity is checked run by run,
the two halves are compared as means of per-run scales, and the
bootstrap of the joint solution is retained only as a diagnostic. Under
this version all fifty surveys of set B pass, and the drift of the time
offset along the survey (Section 3.3), the hand-eye residual and the
dispersion of the per-run scales remain the criteria that reject a
defective alignment.

\subsection{Limitations}\label{limitations}

The LiDAR reference is a consumer SLAM instrument whose drift over long
paths has not been characterised independently; its per-point accuracy
does not limit a scale comparison drawn from millions of points, and it
agrees with the smartphone reference to within 1\% on average with no
systematic sign, but a tape measurement of a known object on one of the
LiDAR sites remains the only way to validate either reference without
assuming the other. The results of Section 5.4 were computed with the
joint estimator and could not be recomputed on all twelve models. The
estimator reads one lens of the rig; the two lenses give similar
accuracy against the LiDAR but differ by 0.6\% on average on the
production surveys, for a reason not isolated here. All surveys with
external reference were recorded with the X5, while the X6 is
characterised as a sensor only. The static characterisation was
performed on a third camera unit and establishes orders of magnitude
rather than unit-specific figures; a longer static recording would allow
a proper Allan-variance analysis of bias instability. Finally, the
linear alignment is not followed by a non-linear refinement with
preintegration factors and gyroscope bias, which the visual-inertial
SLAM literature uses to improve on the closed-form initialisation.

\section{Conclusion}\label{conclusion}

The IMU embedded in the Insta360 X5 and X6, combined with the camera
poses that the SfM pipeline already computes, yields a metric scale
without smartphone data. The sensor is recorded at 1 kHz on the video
clock to within \textpm{}5 ms, its accelerometer scale factor is correct
to 0.03\% and its bias moves by 0.042 m\,s$^{-2}$ over ten minutes. A
linear visual-inertial alignment recovers the scale of existing
reconstructions, independently of the units and orientation of the input
model, provided that the frame timing is estimated from the data rather
than taken at its nominal value and that the alignment is solved per
contiguous segment rather than jointly over the survey. The segment-wise
estimate reduces the mean absolute error against external reference from
3.43\% to 0.85\%. On nine surveys with a co-mounted LiDAR the inertial
scale deviates by 0.81\% in mean absolute error against 1.94\% for the
smartphone reference, and on printed fiducial markers both references
fall within the 1\% uncertainty of the target. On the reference
trajectories of thirty sequences of a public benchmark the estimator is
unbiased, with a mean error of $-$0.07\% and a dispersion of 1.28\% per
sequence, whatever the scale of the input. On fifty production surveys
the two references agree to a standard deviation of 2.3\%, with an
unattributed mean difference of 0.3\% that varies with the extent of the
smartphone anchor and, by a similar amount, with the lens of the rig
used for the estimate. The smartphone capture step can therefore be
removed from the procedure, with the recovered gravity magnitude, the
drift of the video-to-IMU offset and the dispersion of the per-run
scales available to reject a failed alignment. Open items are a third
metric reference on the production sample, which is the only way to
assign the residual difference to one of the two estimates, and a
non-linear refinement of the linear solution.

\end{document}